\documentclass{webofc}

\usepackage[varg]{txfonts}   
\usepackage{hyperref}
\usepackage{graphicx}
\usepackage{algorithm2e}
\usepackage{multirow}
\usepackage[caption=false]{subfig}
\usepackage{xcolor}
\usepackage{amsmath, amssymb}
\usepackage{tabularx}
\usepackage{comment}
\usepackage{threeparttable} 
\usepackage{array}          

\hypersetup{colorlinks=true,citecolor=blue,urlcolor=blue,linkcolor=blue}

\begin{document}

\title{HPCNeuroNet: A Neuromorphic Approach Merging SNN Temporal Dynamics with Transformer Attention for FPGA-based Particle Physics}

\author{
    \firstname{Murat} \lastname{Isik}\inst{1} \and
    \firstname{Hiruna} \lastname{Vishwamith}\inst{2} \and
    \firstname{Jonathan} \lastname{Naoukin}\inst{3} \and
    \firstname{I. Can} \lastname{Dikmen}\inst{4}
}

\institute{
    Stanford University, Stanford, USA \and
    University of Moratuwa, Moratuwa, Sri Lanka \and
    University of Texas at Austin, Austin, USA \and
    Temsa R\&D Center, Adana, Turkey
}

\abstract{This paper presents the innovative HPCNeuroNet model, a pioneering fusion of Spiking Neural Networks (SNNs), Transformers, and high-performance computing tailored for particle physics, particularly in particle identification from detector responses. Our approach leverages SNNs' intrinsic temporal dynamics and Transformers' robust attention mechanisms to enhance performance when discerning intricate particle interactions. At the heart of HPCNeuroNet lies the integration of the sequential dynamism inherent in SNNs with the context-aware attention capabilities of Transformers, enabling the model to precisely decode and interpret complex detector data. HPCNeuroNet is realized through the HLS4ML framework and optimized for deployment in FPGA environments. The model accuracy and scalability are also enhanced by this architectural choice. Benchmarked against machine learning models, HPCNeuroNet showcases better performance metrics, underlining its transformative potential in high-energy physics. We demonstrate that the combination of SNNs, Transformers, and FPGA-based high-performance computing in particle physics signifies a significant step forward and provides a strong foundation for future research.
}

\maketitle

\vspace{-25pt}

\section{Introduction}
\label{intro}
Particle physics faces challenges in particle identification and hardware limitations. ML techniques have made strides, but neuromorphic computing offers a transformative, energy-efficient solution. Neuromorphic Computing, blending biological systems with artificial intelligence, has emerged as a transformative solution, surpassing traditional computing paradigms by up to 100,000 times in efficiency \cite{isik2022design, isik2023survey}. Hybrid models such as SpikeFormer, SSTFormer, and SpikeGPT combine the strengths of deep learning and energy-efficient spiking systems, offering solutions to computational challenges \cite{li2022spikeformer, zhu2023spikegpt, wang2023sstformer}. Enabling these state-of-the-art algorithms is the HLS4ML package, which leverages High-Level Synthesis (HLS) to deliver accelerated and efficient solutions \cite{fahim2021hls4ml, WinNT8, iiyama2021distance}. The flexibility and parallel processing capabilities of FPGAs have made them a computational backbone in modern particle physics research.

\vspace{-2pt}

\begin{figure}[h]
    \centering
    \includegraphics[width = 0.35\textwidth]{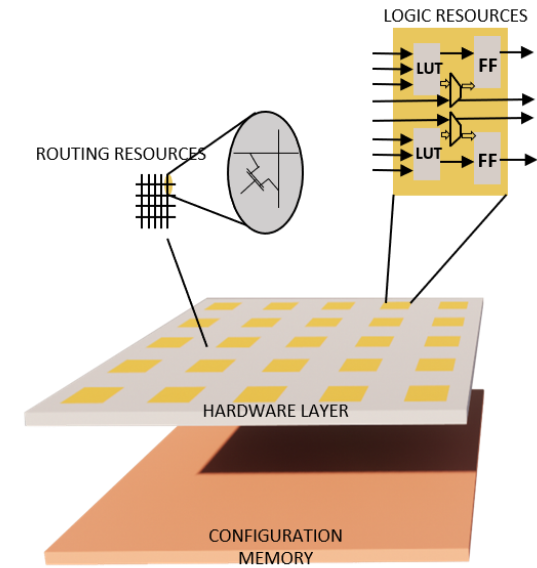}
        \caption{The fundamental FPGA architecture.}
        \label{FPGA}
\end{figure}

The low-latency operation of FPGAs further capacitates real-time data analysis, a significant boon in the fast-paced realm of particle physics. Coupled with tools like HLS4ML that bridge high-level programming constructs with FPGA-compatible descriptions, the process of algorithm deployment becomes streamlined. As research delves deeper into the intersection of neuromorphic computing and particle physics, the flexibility and efficiency of FPGAs stand out as a vital asset in this interdisciplinary nexus. In the evolving landscape of particle physics research, the computational backbone has increasingly leaned toward the capabilities of FPGAs \cite{duarte2019fpga, que2022optimizing, barbosa2023development, duarte2022efficient}. We propose a design methodology for HPCNeuroNet, combining SNN temporal dynamics with Transformer attention mechanisms optimized for FPGA-based particle physics. The methodology includes:

\begin{itemize}
    \item An efficient hybrid neural network that integrates SNN dynamics and Transformer attention, ensuring resilience and adaptability for complex computations.
    
    \item Robust attention mechanisms to effectively process particle collision data, enhancing tolerance to discrepancies and anomalies.
    
    \item We advance a system software that partitions the model into clusters, enabling resource-efficient and energy-optimized FPGA implementation.
\end{itemize}

This paper presents HPCNeuroNet, a novel high-performance computing approach that combines neuromorphic computing with transformer technology to address particle detection challenges. \textbf{Section 2} provides background on particle physics data processing, emphasizing advancements in neuromorphic computing for FPGA and the synergy between SNNs and Transformer attention mechanisms. \textbf{Section 3} introduces the core methodology of HPCNeuroNet, optimized for FPGA deployment. \textbf{Section 4} presents performance results and benchmarks. \textbf{Section 5} concludes the study by summarizing key contributions, while \textbf{Section 6} outlines directions for future research.

\vspace{-10pt}

\section{Neural Networks in Particle Physics}
\begin{figure}[h]
    \centering
    \includegraphics[width = 0.7\textwidth]{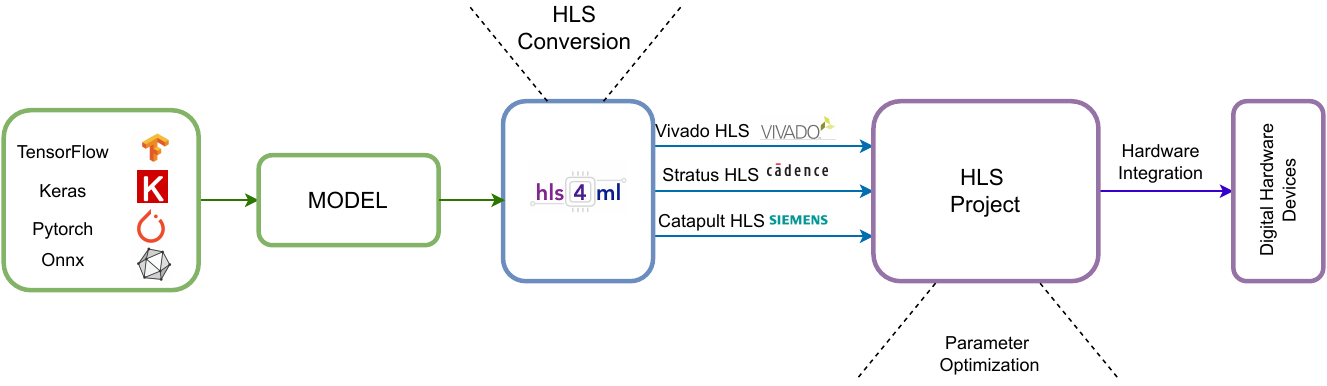}
        \caption{A standard procedure to convert an ML model for digital hardware devices deployment via hls4ml is depicted. The model training and framework, represented by the green boxes (left), are executed in traditional ML software frameworks. The hls4ml setup and transformation processes are demonstrated in the blue boxes (center). The purple boxes (right) highlight potential methods to incorporate the HLS project into a more extensive hardware design.}
        \label{roadmap2}
\end{figure}

\vspace{-2pt}
The incorporation of ML techniques has brought transformative advancements to Particle Physics. Graph Neural Networks (GNNs) represent particle collisions in a geometric framework, where nodes encapsulate particle properties \cite{duarte2022graph, butter2023machine}. In contrast, Convolutional Neural Networks (CNNs) excel at pattern recognition, while Classification Algorithms significantly enhance data categorization efficiency, replacing manual methods \cite{uboldi2022extracting, al2022cnn, schwartz2021modern, lu2021sparse}. However, these approaches remain limited by classical computer architectures and Moore’s Law, which restricts computational power. To transcend these limitations, neuromorphic computing emerges as a leading solution alongside Quantum Computers, Photonic Computers, and AI-specific hardware like Tensor Processing Units \cite{huynh2022implementing, kosters2023benchmarking}. Hybrid ML models, such as combining CNNs with Principal Component Analysis, address the complexities of subatomic interactions and massive datasets, particularly in large hadron collider experiments. These models offer robust and efficient solutions where conventional methods fall short. Inspired by the brain’s neural processes, neuromorphic systems enable rapid parallel processing and unmatched energy efficiency, revolutionizing particle interaction analysis \cite{r2023sensor, strukov2019building}. HLS4ML bridges this gap by converting complex ML models into firmware for deployment on hardware platforms like FPGAs \cite{fahim2021hls4ml}. With reconfigurability and parallel processing capabilities, FPGAs enable rapid computations with minimal latency, creating a synergistic environment for hybrid models. The integration of HLS4ML with FPGA has enabled efficient, real-time, and scalable particle identification techniques, as shown in \autoref{roadmap2}. While established methodologies have laid the foundation for current advancements, they also revealed limitations. Advanced FPGA platforms, combined with hybrid models, represent a transformative shift, offering unprecedented precision and clarity in particle physics \cite{khoda2023ultra, ghielmetti2022real}.

\vspace{-30pt}

\section{Proposed Design Methodology}
\label{sec-2}

\subsection{Datasets}

\subsubsection{CERN CMS Electron Collision Data}  

This dataset contains 100k dielectron events within the invariant mass range of 2–110 GeV, selected for outreach and education purposes, and represents a subset of CMS physics results \cite{WinNT1}. The momentum components of Electron 1—first, second, and third (in GeV)—exhibit Gaussian distributions, while the transverse momentum follows an exponential decay due to the lack of symmetry elements. The second electron’s characteristics show similar trends. Notably, the phi angles for both electrons display linear behavior, while the transverse momentum for both follows an exponential decay. Gaussian functions effectively describe the behavior of the first three momentum components, while transverse momentum reflects an exponential decay trend. Upon inspection of the Electron 1 counts, the first, second, and third momentum components (expressed in units of GeV) predominantly showcase Gaussian distributions.

\vspace{-10pt}

\subsubsection{CERN DUNE Particle Identification from Detector Responses}

This dataset identifies Positrons, Pions, Kaons, and Protons, with Pions, Kaons, and Protons being the most frequent, while Positrons are rarer due to their antiparticle nature \cite{WinNT2}. The momentum follows a Gaussian distribution between 0.21 GeV/c and 5.29 GeV/c, with the most common range being 0.21–1.33 GeV/c. The theta angle peaks at 0.26–0.35 radians with 400,000–460,000 counts. The beta values form a half-Gaussian shape, rising between 0.55–0.99, peaking at 0.99–1.01 with ~918,000 counts, and then sharply dropping between 1.01–1.02 with rare counts extending to 1.5. Photoelectron counts mostly range from 0 to 7, with fewer outliers up to 349 photoelectron. Energy values show similar trends, peaking at 0–0.02 GeV and declining rapidly. The inner energy decreases continuously, while the outer energy shows a slight increase between 0.04–0.07 GeV before tapering off (0.9 GeV for inner energy, 1.1 GeV for outer energy).

\vspace{-10pt}

\subsubsection{CERN CMS Proton Collision Data Observations and Count Models}

This dataset includes 20,000 proton collision events from lumi section 388-1804, curated for outreach and education \cite{WinNT3}. The majority of detections occur in lumi sections 954–1,025, with most events clustering around the 1 billion mark (~8,000 counts). Additional distributions range from 302 million to 512.5 million and 1.63 billion to 1.70 billion, with 4,000–4,700 counts. Unlike the Electron Collision data, this dataset predominantly exhibits exponential decay rather than Gaussian functions. Exponential trends are observed in the overall mass scale, ratio squared, and components of the leading and subleading megajet vectors. The number of jets with transverse momentum exceeding 40 GeV also follows a similar decay pattern.

\vspace{-10pt}

\subsection{HPCNeuroNet}

HPCNeuroNet is a state-of-the-art architecture tailored for the efficient processing of time-series data, leveraging the power of the HLS4ML library for high-level synthesis. This framework has been tested and optimized for various datasets. The raw time-series data feeds directly into the \textit{Transformer Embedding} layer, where it's transformed into dense vector embeddings. These embeddings encapsulate the essential features of the time-series data, preparing them for processing by the subsequent Transformer layers. The core computational layers, the \textit{HPCNeuroNet Transformer Layers}, process these embeddings to discern patterns inherent to the time-series data. Within these layers, the \textit{Self-Attention Mechanism} plays a pivotal role. It uses transformations into Query (Q), Key (K), and Value (V) matrices, facilitating processes like \textit{Scaled Dot-Product Attention}, \textit{Multi-Head Attention}, and \textit{Positional Encoding}. Following the Transformer layers, the data transitions to the \textit{Spiking Self Attention} mechanism, which introduces the dynamics of SNNs. The processed data is then encoded via convolutional layers in the \textit{SNN Encode} stage and subsequently decoded through linear layers in the \textit{SNN Decode} stage. The entire process culminates in the generation of a refined output, representing the enhanced and processed time-series data, ready for further analysis, and application in desired domains.

\vspace{-10pt}

\subsection{Experimental Setup}

\subsubsection{CPU/GPU Implementation}

Our algorithms were implemented using Python on CPUs and GPUs. The study was carried out by leveraging the computational power of NVIDIA's GeForce RTX 3060 GPU and Intel's Core i9 12900H CPU, both of which are optimized for different tasks so that our implementations were executed efficiently.

\subsubsection{FPGA Implementation}

\begin{figure*}
    \graphicspath{ {D:\Stack} }
    \center \includegraphics[width=0.7\textwidth]{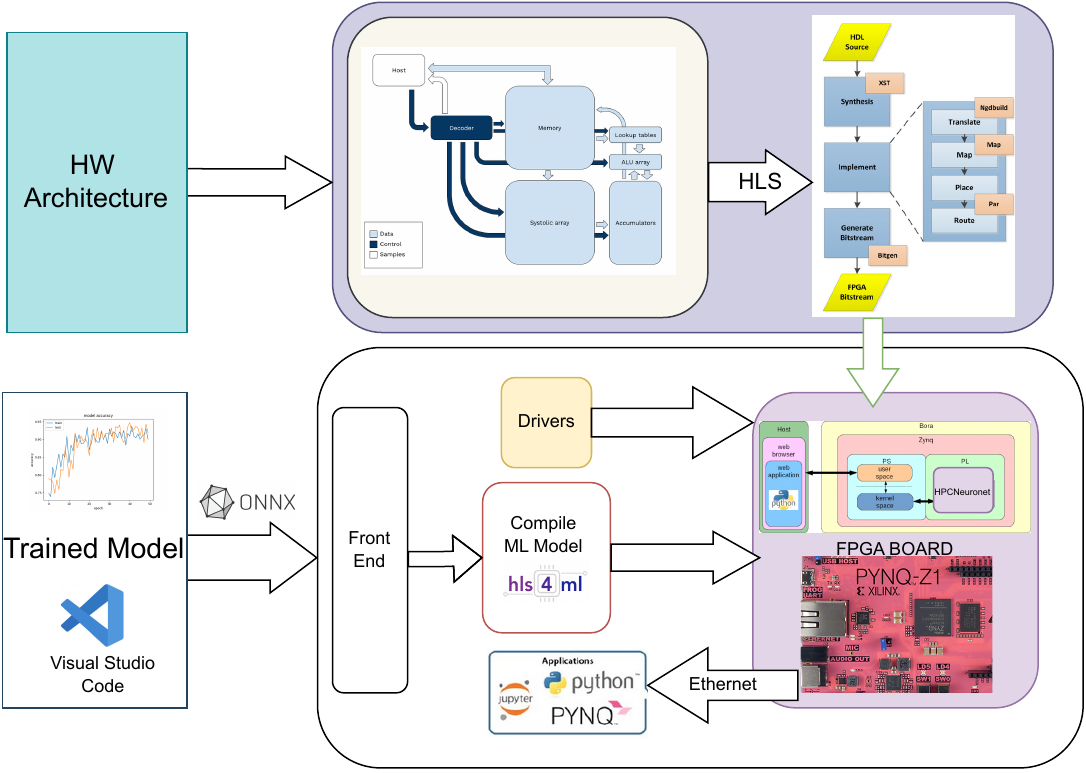}
    \caption{Block Diagram of Implementation}
    \label{Figure 3}
\end{figure*}

This section explains the procedure to implement the HPCNeuroNet, a hybrid model comprising of both Transformer and SNN components, on an FPGA using the HLS4ML library and the PYNQ framework, as illustrated in \autoref{Figure 3}. The process begins with setting up the PYNQ framework on an FPGA board (PYNQ-Z1) using a pre-built or custom image. A pre-trained HPCNeuroNet model, trained in TensorFlow or Keras, is then converted into HLS code using HLS4ML, with configurations specifying precision, layer types, and optimizations. The HLS code is synthesized into RTL designs via Vivado HLS, and a bitstream file (.bit) is generated, transferred to the PYNQ board, and programmed using the PYNQ Python API. Once deployed, inferences are executed, and performance metrics such as latency, throughput, and accuracy are analyzed.

\vspace{-15pt}

\section{Evaluation}
\label{sec-3}

\vspace{-5pt}

The results in \autoref{table:table_2} provide a comparative analysis of HPCNeuroNet’s accuracy, MAC operations, and latency across different models. The CMS-Electron Collision model achieves the highest accuracy at 94.48\%, requiring the fewest computational operations (0.49 GOP) and the lowest latency (11.5 ms), making it both precise and efficient. In contrast, the CMS-Proton Collision model delivers intermediate accuracy at 88.73\%, with a latency of 13.5 ms and higher MAC operations (1.29 GOP), reflecting its greater complexity. The Geant4-Particle Identification model, though achieving a lower accuracy of 78.05\%, demonstrates commendable performance given the task’s complexity, requiring 1.09 GOP and a latency of 13.1 ms. While the differences in latency and accuracy appear minor, they can have significant implications for real-time applications. Notably, the CMS-Electron Collision model focuses on regression, whereas the others address classification tasks, influencing the interpretation of accuracy metrics.

\begin{table}[h]
\scriptsize
\centering
\caption{Evaluation results for CERN Electron Collison, LArTPC waveforms, and Particle Identification from Detector Responses on HPCNeuronet.}
\label{table:table_2}
\begin{tabular}{c|c|c|c|c}
\hline
& \textbf{Accuracy} & \textbf{MAC (GOP)} & \textbf{Task} & \textbf{Latency (ms)} \\ 
\hline
\textbf{CMS-Electron Collison \cite{WinNT1}} & 94.48\% & 0.49 &Regression & 11.5 \\ 
\hline
\textbf{Geant4-Particle Identification from Detector Responses \cite{WinNT2}} & 78.05\% & 1.09 & Classification &  13.1\\ 
\hline
\textbf{CMS-Proton Collision
\cite{WinNT3}} & 88.73\% & 1.29 & Classification&  13.5\\ 
\hline

\end{tabular}
\label{tab:comparison}
\end{table}

\begin{figure*}%
    \centering
    \begin{minipage}{0.4\textwidth}
        \centering
        \subfloat[Latency comparison.\label{fig:original_connection}]{{\includegraphics[width=0.95\linewidth]{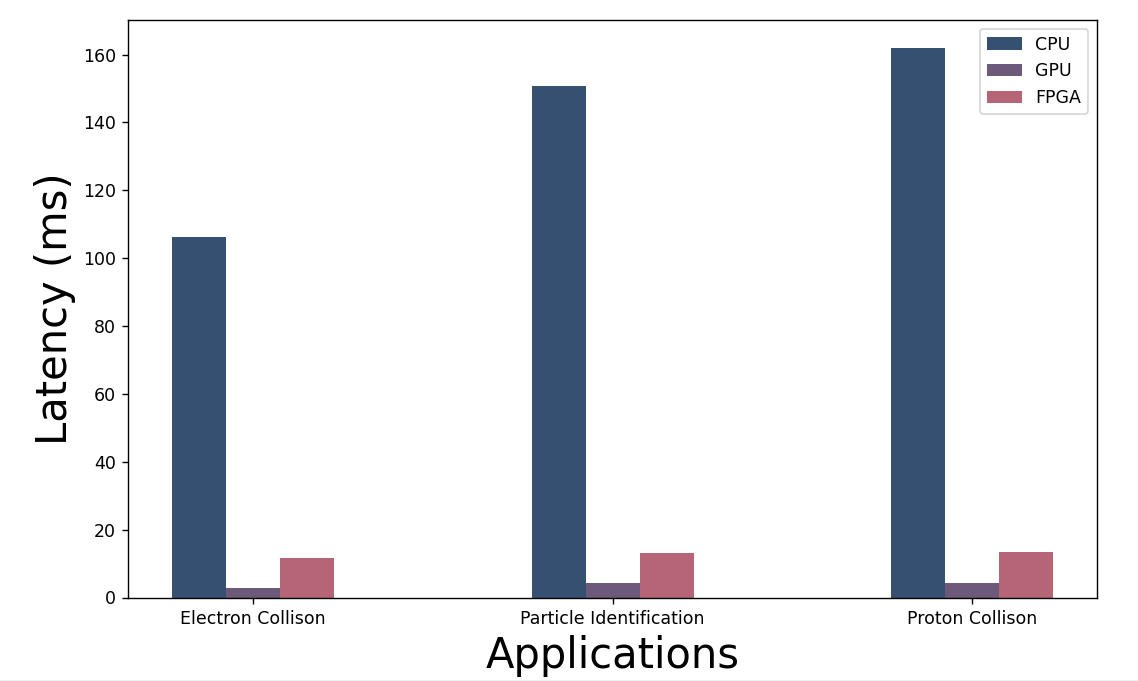} }}
    \end{minipage}%
    \begin{minipage}{0.4\textwidth}
        \centering
        \subfloat[Power Consumption comparison.\label{fig:astrocyte_modulation1}]{{\includegraphics[width=0.95\linewidth]{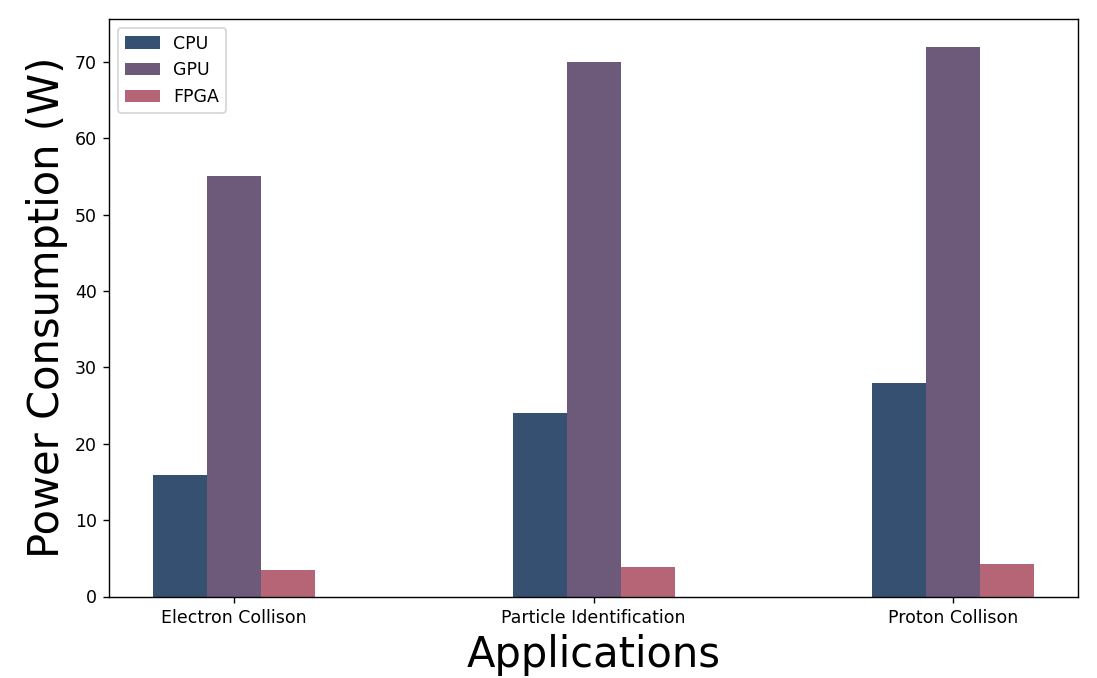} }}
    \end{minipage}
    
    \vspace{1em} 
    
    \begin{minipage}{0.4\textwidth}
        \centering
        \subfloat[Throughput comparison.\label{fig:astrocyte_modulation2}]{{\includegraphics[width=0.95\linewidth]{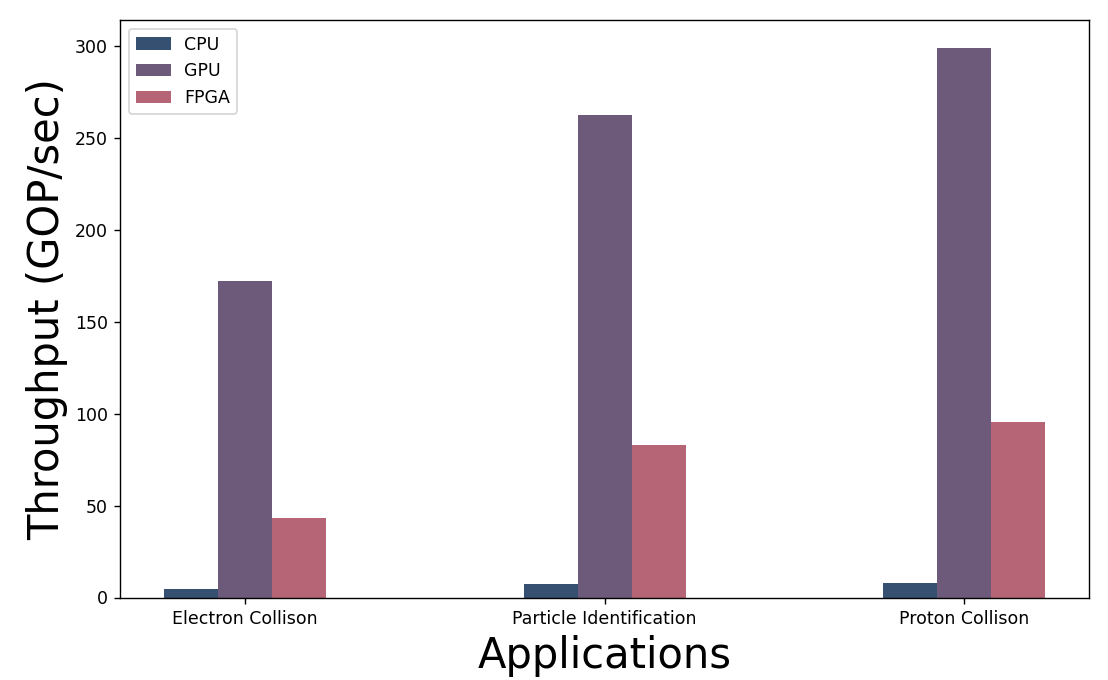} }}
    \end{minipage}%
    \begin{minipage}{0.4\textwidth}
        \centering
        \subfloat[Power Efficiency comparison.\label{fig:astrocyte_modulation3}]{{\includegraphics[width=0.95\linewidth]{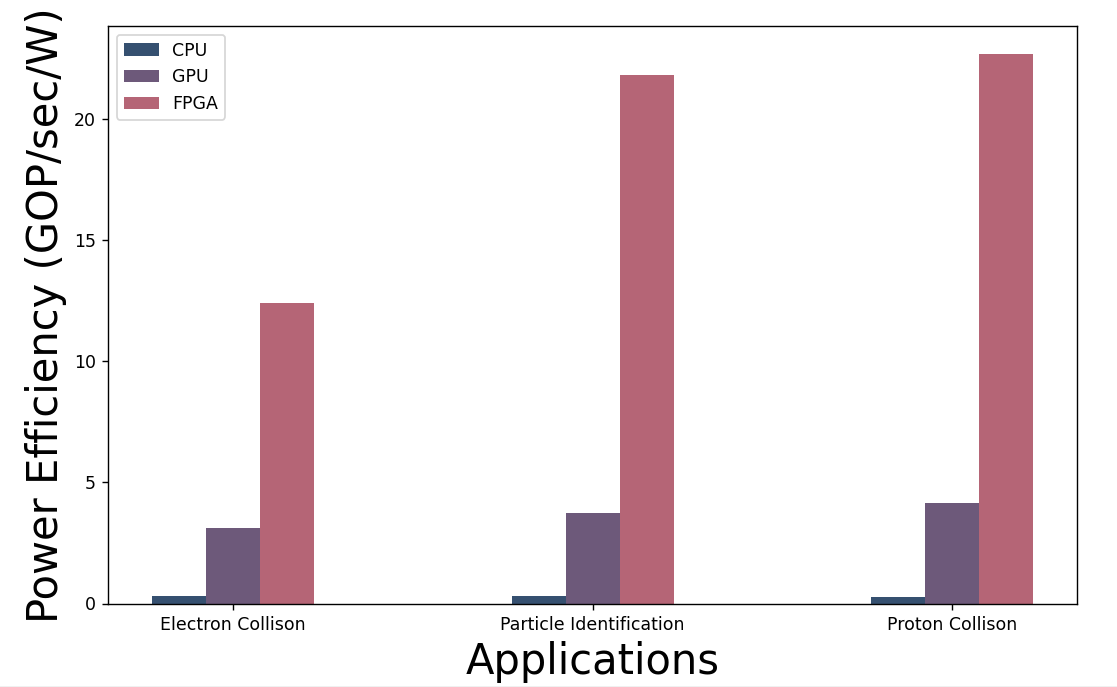} }}
    \end{minipage}
    
\caption{Performance of HPCNeuronet across three distinct hardware platforms using various datasets.}
    \label{fig:res_neural_network}
\end{figure*}

\autoref{fig:res_neural_network} illustrates a number of key metrics, such as manufacturing technology, operating frequency, power consumption, as well as latency, throughput, and power efficiency. There is evidence of this when examining the latency metric for the Xilinx FPGA, which stands at a low value. With regard to power consumption, the FPGA demonstrates remarkable energy efficiency with a power requirement significantly lower than that of both the CPU and GPU. Moreover, the figure shows the performance-per-watt of the FPGA with a high throughput and a high level of power, underscoring their suitability for tasks where power efficiency is a critical factor and their use as hardware accelerators.

\begin{table}[h]
    \caption{Comparison of our HPCNeuroNet framework with other models.}
    \renewcommand{\arraystretch}{1}
    \setlength{\tabcolsep}{2pt}
    \scriptsize
    \centering
    \begin{threeparttable}
    {\fontsize{9}{10}\selectfont
        \begin{tabular}{c|cccc}
            \hline
            Models & DNN & GNN & 1-D CNN & \textbf{HPCNeuronet} \\
            & & & & \small{\textbf{(Ours)}}\\
            \hline
            
            Accuracy (\%) & 85.1 & 78.3 & 85.8 & \textbf{88.73} \\
            MAC (GOP) & 1.32 & 2.12 & 0.77 & \textbf{1.29} \\
            Latency [ms] & 24.2 & 32.9 & 18.3 & \textbf{11.5} \\
            \hline
            Power Efficiency &10.4  & 3.5 & 12.2 & \textbf{22.7} \\
            \hline
        \end{tabular}}
        \begin{tablenotes}[flushleft]
            \item *Power Efficiency unit is GOP/s/W.
            \item *CMS-Electron Collison dataset was implemented on these models.
        \end{tablenotes}
    \end{threeparttable}
    
    \label{tab:comparison_others}
\end{table}

\vspace{-10pt}

\autoref{tab:comparison_others} compares the performance metrics of HPCNeuroNet with DNN, GNN, and 1-D CNN models, each tailored to specific applications. HPCNeuroNet achieves the highest accuracy at 88.73\%, showcasing its robustness and efficiency. Despite its superior accuracy, HPCNeuroNet also delivers low latency at 11.5 ms, significantly outperforming DNN and GNN models, thanks to design optimizations prioritizing real-time performance. The model’s MAC operations, measured at 1.29 GOP, reflect its computational efficiency, especially given its accuracy. Additionally, HPCNeuroNet achieves exceptional power efficiency of 22.7, far surpassing the compared models, making it ideal for power-sensitive applications.

\vspace{-15pt}

\section{Conclusions}
\label{sec-4}

\vspace{-5pt}

The proposed HPCNeuroNet model integrates Spiking SNN temporal dynamics with Transformer attention mechanisms, excelling at complex particle identification tasks. Benchmarks demonstrate significant improvements in performance, accuracy, and computational speed, particularly with FPGA deployment using the HLS4ML framework. This integration also ensures energy efficiency, a critical factor for large-scale particle detection experiments. While implementation challenges exist, such as software version compatibility and FPGA-specific nuances, the results highlight HPCNeuroNet's transformative potential in high-energy physics. Beyond particle physics, this fusion of SNNs, Transformers, and FPGA-based computing opens avenues for broader scientific applications, showcasing the power of neuromorphic computing in advancing research domains.

\vspace{-10pt}

\section{Future Work}
\label{sec-5}

\vspace{-5pt}

Our research demonstrates the potential of FPGA-based accelerators in particle physics and outlines future directions. Enhancing SNN dynamics, extending HPCNeuroNet with diverse models or attention mechanisms, and leveraging Dynamic Partial Reconfiguration (DPR) can improve adaptability and efficiency. Exploring photonic and neuromorphic computing could reveal complementary strengths, while the robustness of neuromorphic systems makes them ideal for pattern recognition tasks.

\vspace{-10pt}

\section{Acknowledgements}

\vspace{-5pt}

We acknowledge the Temsa Research R\&D Center for their generous financial support and the reviewers for their invaluable insights and suggestions that significantly contributed to the enhancement of our paper.

\vspace{-10pt}
\bibliography{external}

\begin{thebibliography}{27}

\bibitem{isik2022design}
M.~Isik, A.~Paul, M.L. Varshika, A.~Das, A design methodology for fault-tolerant computing using astrocyte neural networks, in \emph{Proceedings of the 19th ACM International Conference on Computing Frontiers} (2022), pp. 169--172

\bibitem{isik2023survey}
M.~Isik, A survey of spiking neural network accelerator on fpga, arXiv preprint arXiv:2307.03910  (2023).

\bibitem{li2022spikeformer}
Y.~Li, Y.~Lei, X.~Yang, Spikeformer: A novel architecture for training high-performance low-latency spiking neural network, arXiv preprint arXiv:2211.10686  (2022).

\bibitem{zhu2023spikegpt}
R.J. Zhu, Q.~Zhao, J.K. Eshraghian, Spikegpt: Generative pre-trained language model with spiking neural networks, arXiv preprint arXiv:2302.13939  (2023).

\bibitem{wang2023sstformer}
X.~Wang, Z.~Wu, Y.~Rong, L.~Zhu, B.~Jiang, J.~Tang, Y.~Tian, Sstformer: Bridging spiking neural network and memory support transformer for frame-event based recognition, arXiv preprint arXiv:2308.04369  (2023).

\bibitem{fahim2021hls4ml}
F.~Fahim, B.~Hawks, C.~Herwig, J.~Hirschauer, S.~Jindariani, N.~Tran, L.P. Carloni, G.~Di~Guglielmo, P.~Harris, J.~Krupa et~al., hls4ml: An open-source codesign workflow to empower scientific low-power machine learning devices, arXiv preprint arXiv:2103.05579  (2021).

\bibitem{WinNT8}
{HLS4ML}, \url{https://fastmachinelearning.org/}, accessed: 2023-10-18

\bibitem{iiyama2021distance}
Y.~Iiyama, G.~Cerminara, A.~Gupta, J.~Kieseler, V.~Loncar, M.~Pierini, S.R. Qasim, M.~Rieger, S.~Summers, G.~Van~Onsem et~al., Distance-weighted graph neural networks on fpgas for real-time particle reconstruction in high energy physics, Frontiers in big Data \textbf{3}, 598927 (2021).

\bibitem{duarte2019fpga}
J.~Duarte, P.~Harris, S.~Hauck, B.~Holzman, S.C. Hsu, S.~Jindariani, S.~Khan, B.~Kreis, B.~Lee, M.~Liu et~al., Fpga-accelerated machine learning inference as a service for particle physics computing, Computing and Software for Big Science \textbf{3}, 1 (2019).

\bibitem{que2022optimizing}
Z.~Que, M.~Loo, H.~Fan, M.~Pierini, A.~Tapper, W.~Luk, Optimizing Graph Neural Networks for Jet Tagging in Particle Physics on FPGAs, in \emph{2022 32nd International Conference on Field-Programmable Logic and Applications (FPL)} (IEEE, 2022), pp. 327--333

\bibitem{barbosa2023development}
F.~Barbosa, L.~Belfore, N.~Branson, C.~Dickover, C.~Fanelli, D.~Furletov, S.~Furletov, L.~Jokhovets, D.~Lawrence, D.~Romanov, Development of ml fpga filter for particle identification and tracking in real time, IEEE Transactions on Nuclear Science  (2023).

\bibitem{duarte2022efficient}
J.~Duarte, M.~Liu, J.~Ngadiuba, E.~Cuoco, J.~Thaler, Efficient ai in particle physics and astrophysics, Frontiers in Artificial Intelligence \textbf{5}, 999173 (2022).

\bibitem{duarte2022graph}
J.~Duarte, J.R. Vlimant, in \emph{Artificial intelligence for high energy physics} (World Scientific, 2022), pp. 387--436

\bibitem{butter2023machine}
A.~Butter, T.~Plehn, S.~Schumann, S.~Badger, S.~Caron, K.~Cranmer, F.A. Di~Bello, E.~Dreyer, S.~Forte, S.~Ganguly et~al., Machine learning and lhc event generation, SciPost Physics \textbf{14}, 079 (2023).

\bibitem{uboldi2022extracting}
L.~Uboldi, D.~Ruth, M.~Andrews, M.H. Wang, H.J. Wenzel, W.~Wu, T.~Yang, Extracting low energy signals from raw lartpc waveforms using deep learning techniques—a proof of concept, Nuclear Instruments and Methods in Physics Research Section A: Accelerators, Spectrometers, Detectors and Associated Equipment \textbf{1028}, 166371 (2022).

\bibitem{al2022cnn}
A.~Al-Zoubi, G.~Martino, F.H. Bahnsen, J.~Zhu, H.~Schlarb, G.~Fey, CNN Implementation and Analysis on Xilinx Versal ACAP at European XFEL, in \emph{2022 IEEE 35th International System-on-Chip Conference (SOCC)} (IEEE, 2022), pp. 1--6

\bibitem{schwartz2021modern}
M.D. Schwartz, Modern machine learning and particle physics, arXiv preprint arXiv:2103.12226  (2021).

\bibitem{lu2021sparse}
Y.~Lu, J.~Collado, D.~Whiteson, P.~Baldi, Sparse autoregressive models for scalable generation of sparse images in particle physics, Physical Review D \textbf{103}, 036012 (2021).

\bibitem{huynh2022implementing}
P.K. Huynh, M.L. Varshika, A.~Paul, M.~Isik, A.~Balaji, A.~Das, Implementing spiking neural networks on neuromorphic architectures: A review, arXiv preprint arXiv:2202.08897  (2022).

\bibitem{kosters2023benchmarking}
D.J. K{\"o}sters, B.A. Kortman, I.~Boybat, E.~Ferro, S.~Dolas, R.~Ruiz~de Austri, J.~Kwisthout, H.~Hilgenkamp, T.~Rasing, H.~Riel et~al., Benchmarking energy consumption and latency for neuromorphic computing in condensed matter and particle physics, APL Machine Learning \textbf{1} (2023).

\bibitem{r2023sensor}
S.~R.~Kulkarni, A.~Young, P.~Date, N.~Rao~Miniskar, J.~Vetter, F.~Fahim, B.~Parpillon, J.~Dickinson, N.~Tran, J.~Yoo et~al., On-Sensor Data Filtering using Neuromorphic Computing for High Energy Physics Experiments, in \emph{Proceedings of the 2023 International Conference on Neuromorphic Systems} (2023), pp. 1--8

\bibitem{strukov2019building}
D.~Strukov, G.~Indiveri, J.~Grollier, S.~Fusi, Building brain-inspired computing, Nature Communications pp. 4838--2019 (2019).

\bibitem{khoda2023ultra}
E.E. Khoda, D.~Rankin, R.T. de~Lima, P.~Harris, S.~Hauck, S.C. Hsu, M.~Kagan, V.~Loncar, C.~Paikara, R.~Rao et~al., Ultra-low latency recurrent neural network inference on fpgas for physics applications with hls4ml, Machine Learning: Science and Technology \textbf{4}, 025004 (2023).

\bibitem{ghielmetti2022real}
N.~Ghielmetti, V.~Loncar, M.~Pierini, M.~Roed, S.~Summers, T.~Aarrestad, C.~Petersson, H.~Linander, J.~Ngadiuba, K.~Lin et~al., Real-time semantic segmentation on fpgas for autonomous vehicles with hls4ml, Machine Learning: Science and Technology \textbf{3}, 045011 (2022).

\bibitem{WinNT1}
{CERN Electron Collision Data}, \url{http://opendata.cern.ch/record/304}, accessed: 2023-10-14

\bibitem{WinNT2}
{Particle Identification from Detector Responses }, \url{https://devpost.com/software/geant4-particle-identification-using-ml-algorithms}, accessed: 2022-12-17

\bibitem{WinNT3}
{CERN Proton Collision }, \url{http://opendata.cern.ch/record/1000}, accessed: 2023-10-15

\end{thebibliography}

\end{document}